\documentclass[letterpaper, 10 pt, conference]{ieeeconf}  

\usepackage{amsmath}
\usepackage{amsfonts}
\usepackage{lipsum}
\usepackage{graphicx}
\usepackage{svg}
\usepackage[makeroom]{cancel}

\usepackage{hyperref}
\usepackage[absolute]{textpos}

\usepackage{stfloats}
\usepackage{subcaption}
\usepackage{url}
\usepackage{array} 

\usepackage{xcolor,amsmath}
\usepackage[linesnumbered,ruled,vlined]{algorithm2e}
\DontPrintSemicolon

\SetKwComment{Comment}{\color{green!50!black}// }{}

\newcommand{\xArm}{UFACTORY xArm 6}

\SetKwProg{Function}{function}{}{}

\newcommand\OurMethod{IDA}

\def\KL{D_\textrm{KL}}

\newcommand\stddev[1]{\footnotesize{\textcolor{gray}{#1}}}

\definecolor{tblue}{HTML}{1f77b4}
\definecolor{tgreen}{HTML}{2ca02c}
\definecolor{torange}{HTML}{ff7f0e}
\definecolor{tred}{HTML}{d62728}
\definecolor{tpurple}{HTML}{9467bd}
\definecolor{tgray}{HTML}{7f7f7f}

\IEEEoverridecommandlockouts                              
\title{\LARGE \bf
Information-driven Affordance Discovery \\ for Efficient Robotic Manipulation
}

\author{\authorblockN{\textbf{Pietro Mazzaglia}\textsuperscript{*}}
\authorblockA{Qualcomm AI Research\textsuperscript{\dag}\\
Ghent University\\
}
\and
\authorblockN{\textbf{Taco Cohen}}
\authorblockA{Qualcomm AI Research\textsuperscript{\ddag}}
\and
\authorblockN{\textbf{Daniel Dijkman}}
\authorblockA{Qualcomm AI Research}}

\begin{document}

\begin{textblock}{13}(1.5,0.25)
\centering \noindent\footnotesize © 2024 IEEE.  Personal use of this material is permitted.  Permission from IEEE must be obtained for all other uses, in any current or future media, including reprinting/republishing this material for advertising or promotional purposes, creating new collective works, for resale or redistribution to servers or lists, or reuse of any copyrighted component of this work in other works.
\end{textblock}

\maketitle

\begingroup\renewcommand\thefootnote{*}
\footnotetext{Qualcomm AI Research is an initiative of Qualcomm Technologies, Inc.}
\endgroup
\begingroup\renewcommand\thefootnote{\dag}
\footnotetext{Work done during an internship at Qualcomm AI Research.}
\endgroup
\begingroup\renewcommand\thefootnote{\ddag}
\footnotetext{Work completed while employed at Qualcomm AI Research.}
\endgroup

\thispagestyle{empty}
\pagestyle{empty}

\begin{abstract}
Robotic affordances, providing information about what actions can be taken in a given situation, can aid robotic manipulation. However, learning about affordances requires expensive large annotated datasets of interactions or demonstrations. In this work, we argue that well-directed interactions with the environment can mitigate this problem and propose an information-based measure to augment the agent's objective and accelerate the affordance discovery process. We provide a theoretical justification of our approach and we empirically validate the approach both in simulation and real-world tasks. Our method, which we dub \OurMethod{}, enables the efficient discovery of visual affordances for several action primitives, such as grasping, stacking objects, or opening drawers, strongly improving data efficiency in simulation, and it allows us to learn grasping affordances in a small number of interactions, on a real-world setup with a \xArm\ robot arm.
\end{abstract}
\noindent \textbf{Project website:} \href{https://mazpie.github.io/ida}{mazpie.github.io/ida}

\section{Introduction}

\begin{figure*}[b]
    \centering
    \includegraphics[width=0.75\textwidth]{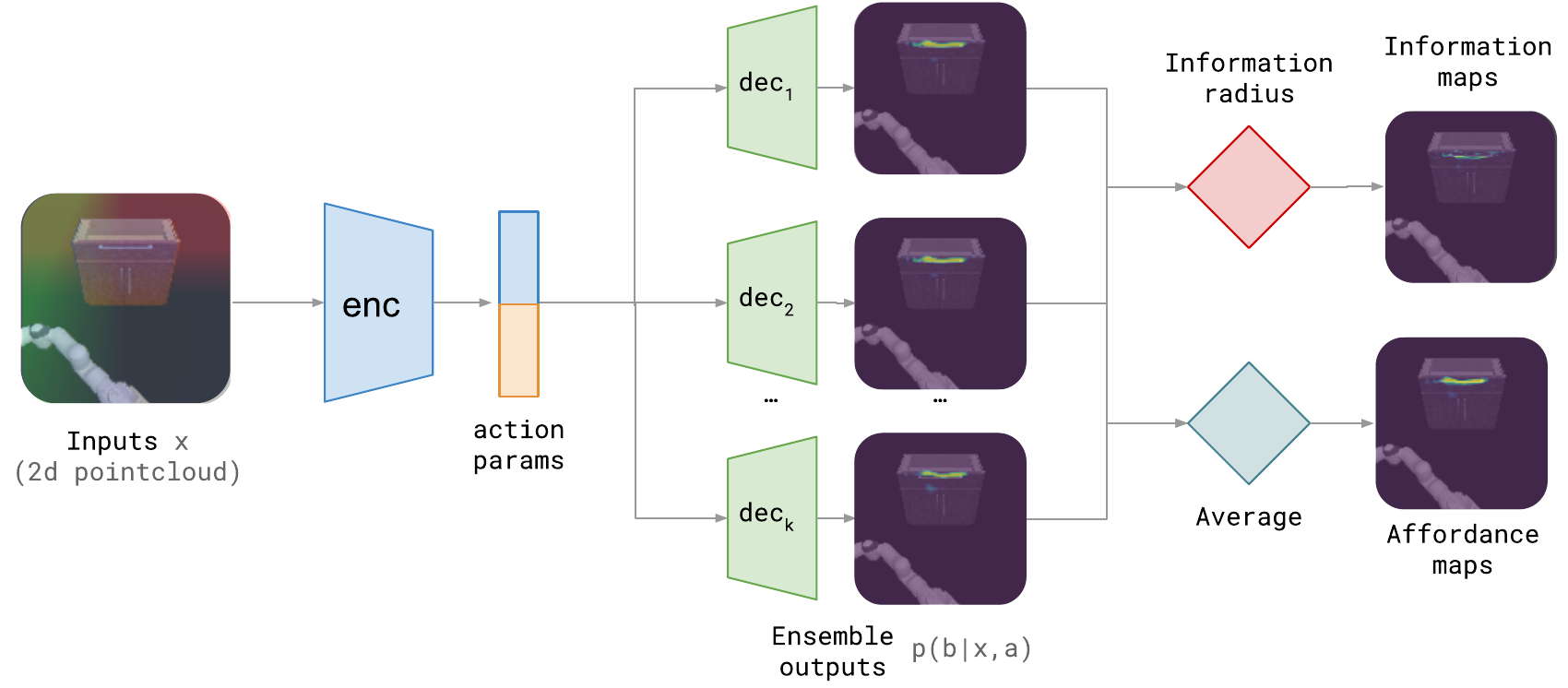}
    \caption{\textbf{Information-driven affordance discovery.} 
    The model processes inputs from the environment (2D point cloud) using a single encoder, concatenates action parameters and decodes visual affordance maps with multiple decoders (ensemble). Averaging the outputs of these networks, we can extract reliable affordance maps, thanks to the ensemble diversity. Computing the information radius, we can obtain information gain maps about affordances in the scene, to drive considerate explorative interactions. Images represent actual model outputs from \OurMethod{} in the ManiSkill2 Open Drawer environment.}
    \label{fig:model}
\end{figure*}

Given the success of learning-based approaches across various domains \cite{Schrittwieser2020MuZero, Dai2021CoatNet, Rombach2022StableDiffusion, Touvron2023LLaMA},  applying learning-based approaches to robotics represents a sensible and appealing idea, in order to develop more general and robust approaches. However, despite recent progress \cite{Ahn2022SayCan, Haarnoja2023SoccerSkills}, learning generalizable robotic systems remains challenging. In particular, robotic manipulation policies can be learned via imitation \cite{Shridhar2022PerAct} or reinforcement learning \cite{Wu2022DayDreamer}, but these struggle to work outside of their training conditions.

In order to produce more generalizable manipulation policies, the robot's perception should provide meaningful representations about how to interact with the environment. \emph{Affordances}, defined as the properties that determine how things in the environment can be used \cite{Gibson2014Affordances}, provide strong cues on how to operate things in the environment. More specifically, visual affordance models generate a visual representation of the environment that maps observed points to a set of possible actions that can be executed at that point \cite{Mo2021Where2Act}. 
However, how to efficiently learn visual affordance models remains a challenge.

Learning visual affordances can be seen as a supervised learning problem, where the agent needs to predict whether a point in the scene is actionable or not, given a certain action primitive. However, if we tackle the problem the classical way, learning an affordance model from a given dataset, training a general and robust model requires large chunks of annotated experiences that are expensive to collect via human interaction or teleoperation \cite{Borjadiaz2022Affordance, Mandikal2021ContactDBAffordances}. Alternatively,
synthetic data can be used to rapidly obtain large datasets to train on \cite{Mo2021Where2Act, Sundermeyer2021ContactGraspnet}, with the risk that the model may not behave as expected in more realistic or complex settings.
In this work, we aim to overcome these limitations, by tackling the affordance discovery problem as an active learning problem. Actively learning what the agent can do in the environment, by continuously interacting with the environment and learning from the new data, should enable faster and more effective learning of affordances, compared to passive data collection from humans or the use of synthetic data \cite{Held1963PassiveKittens}.

We propose \textbf{\OurMethod{}}, an \{I\}nformation-\{D\}riven method for visual \{A\}ffordance discovery, that is able to learn what's feasible in the environment in a small number of interactions, by casting the problem of affordance learning as a contextual bandit problem and exploring actions that are both rich in information and likely to succeed. 

\textbf{Contributions.} Our contributions can be summarized as:
\begin{itemize}
    \item we propose an information-driven measure to augment the agent's objective for visual affordance discovery in interactive settings, and we provide motivation for our approach based on information theory;
    \item we validate \OurMethod{} in simulation, where the agent quickly learns to grasp, stack objects, and open drawers, strongly outperforming previous methods trained on large synthetic datasets \cite{Sundermeyer2021ContactGraspnet}. In this setting, we also show the importance of well-grounded exploration to sensibly improve performance as the number of interactions increases over time;
    \item we demonstrate the applicability of \OurMethod{} in a real robotic setup, with a \xArm, where our agent learns to grasp objects in a small number of interactions, without any prior information.
\end{itemize}

\section{Related Work}

\textbf{Visual robotic manipulation.} Learning to perform manipulation from high-dimensional visual inputs is challenging. Previous work has successfully applied deep learning vision models for robotics, using supervised learning \cite{Levine2016E2EVisuomotor}, reinforcement learning \cite{Kalashnikov2021MTOpt, Lu2021AWOPT}, or world models \cite{Wu2022DayDreamer}. Fully convolutional models, similar to those applied for segmentation \cite{Long2015FCN}, have shown good performance in learning visual tasks such as pushing, grasping \cite{Zeng2018PushGrasp, Ferraro2022OptimRobotics} and rearranging objects \cite{Zeng2022TransporterNet}. U-Net-like models \cite{Ronneberger2015UNet} have been shown to allow efficient reinforcement learning from pixels \cite{James2022QARM}.
The success of fully convolutional models for manipulation could be attributed to the improved capacity of these models to process positional information \cite{Liu2018CoordConv}, which is crucial for manipulation tasks.  

\textbf{Affordance learning.} Previous work on affordances generally focuses on two aspects of the problem: learning the perception modules, that allow detecting affordances in the scene \cite{Mo2021Where2Act,Mo2021O2OAfford}, and learning the action modules \cite{Belkhale2022PLATO, Khazatsky2021ImagineVisAffordances}, which allow translating affordance abstractions into actions. 
One of the closest works to ours is Where2Act \cite{Mo2021Where2Act}, which proposes a perception model and learning-from-interaction framework that combines random and adaptive sampling. The method uses a large number of interactions, simulated through a ``flying" gripper, to learn visual pushing and pulling affordances. To bootstrap learning of the affordance module, random sampling is used until obtaining 10k successful interactions, a procedure that requires sampling hundreds of thousands of trajectories. In contrast, \OurMethod{} achieves greater data efficiency, by actively sampling interactions rich in information.  

\textbf{Exploratory grasping.} 
Grasping remains a challenging problem in robotics \cite{Platt2022GraspLearningReview} due to the high level of complexity required in the interaction between the agent and the objects. Moreover, the large space of potential grasping configurations makes the problem hard to explore. Successful results have been obtained leveraging large synthetic datasets \cite{Mahler2017DexNet2, Mahler2019DexNet4, Sundermeyer2021ContactGraspnet}, but they still struggle to generalize \cite{Wang2019AdversialDexNet}. 
In order to develop more data-efficient approaches, the problem of exploratory grasping \cite{Danielczuk2020ExplGrasp}, i.e. active learning of grasping policies, has been recently studied, leveraging pre-trained models as prior \cite{Li2020ExplWithPrior} or analytical solutions \cite{Fu2022LEGS} to ease the learning process.
Some works also employed a bandit formulation to tackle the problem \cite{Laskey2015MAB, Danielczuk2020ExplGrasp, Lu2020MultifingerActive}.
In our work, we present the idea of actively learning multiple kinds of affordances, including grasping, and present a reliable solution to efficiently learn such affordances by interacting with the environment, without any prior data or knowledge.


\section{Information-driven Affordance Discovery}

\textbf{Setting.} In contextual bandits problems, at each iteration, the agent observes a context {$x \in X$} and selects an action {$a \in A$} to perform, which is rewarded by the environment. In this work, we propose to tackle the affordance discovery problem as a contextual bandit problem. 
The context is the current state of the environment, which is observed by the agent through a camera. For each kind of affordance, e.g., stacking, grasping, opening, actions are represented by parameterizable primitives \cite{Dalal2021RAPS}. For a certain primitive, actions are represented as $a=[p, q]$, where $p$ is the position where the primitive is applied and $q$ is the orientation to apply the primitive.
Rewards represent the affordance success, and so the possible values are $\{0,1\}$.

\textbf{Information gain.} Let $\Theta$ be the space of parameters for an affordance model of the environment, then $p(\theta)$ represents the probability density function of a certain set of parameters to be the best fit for the affordance function. In order to improve our knowledge about what could the best affordance model, we adopt the information gain criterion for exploration, i.e. we want to try actions that maximize the reduction in the entropy with respect to $\Theta$ achieved by learning the outcome of the affordance transition $(x,a,b)$, with outcome $b\in\{\textrm{0 = fail, 1 = success}\}$. The information gained with one transition can be written as:
\begin{equation}
\label{eq:info}
    I(b,x,a) = \KL{}[p(\Theta|b,x,a) \Vert p(\Theta)], 
\end{equation}
that is the Kullback-Leibler divergence between the posterior about the model's parameters given the transition and the prior about the same parameters.

We are interested in measuring the expected amount of information to be gained from performing action $a$ in state $x$. Observing that $p(b,x,a|x,a)=p(b|x,a)$, we can rewrite the above as:
\begin{equation}
\label{eq:info}
    I(x,a) = \mathbb{E}_{p(b|x,a)} [I(b, x, a)].
\end{equation}
In Appendix, we show that this equals the Jensen-Shannon Divergence (JSD) between the distributions $p(b | x, a, \theta)$ for all $\theta$:
\begin{align}
\label{eq:jsd}
    I(x,a) &= JSD(p(b|x,a, \theta)|\theta \sim \Theta) = \\
        &= H(\mathbb{E}_{\theta \sim \Theta}[p(b|x,a, \theta)]) -  \mathbb{E}_{\theta \sim \Theta}[H(p(b|x,a, \theta))], \notag
\end{align}
The JSD, also known as information radius \cite{Sibson1969JSD}, represents the total divergence to the average in a space of distributions. It can also be interpreted as the amount of information that is carried by evidence about the model \cite{Jardine1971MathTaxonomy}. In our context, the JSD captures the amount of information that a context-action pair $(x,a)$ carries about which model parameterization $\theta$ fits the data best. Thus, we can use it as a measure of the utility of the pair in order to improve the affordance model. Crucially, as we also discuss in Section \ref{sec:implementation}, computing the JSD only requires sampling multiple sets of parameters, and thus requires no additional data from the environment or updates of the model.

\textbf{Affordance discovery.}
During the interaction and training stage, we aim to tackle the affordance discovery problem by sampling actions that are informative to learn a better affordance model but also likely to be successful. For the former, we can rely on the JSD presented in Equation \ref{eq:jsd} to measure the information we expect to gain with a new interaction. For the latter, we should consider the probability $p(b|x,a)$ of an affordance to be successful.   

As our setting is analogous to a contextual bandit problem, we can combine our information-driven measure with the expected ``reward" from the environment, by adopting an upper confidence bound (UCB) strategy \cite{Auer2002UCB, Audibert2009UCBVariance}. This implies that the agent samples actions according to an overestimate of the expected reward:
\begin{equation}
\label{eq:ucb}
      \arg \max_{a} [ \hat{r}(x,a) + c_\textrm{expl} \cdot I(x,a) ]
\end{equation}
where the coefficient $c_\textrm{expl} \geq 0$ controls the tradeoff between exploration and exploitation. This means that practically the agent, being in context $x$, should sample actions that satisfy:
\begin{align}
\label{eq:sampling}
    \arg \max_{a} [ p(b|x,a) + c_\textrm{expl} \cdot JSD(p(b|x,a, \theta)|\theta \sim \Theta) ]
\end{align}
that, given that the $JSD$ is non-negative, complies with the upper confidence bound condition of the agent's criterion being greater or equal than $\hat{r}(x,a)$.


\section{Implementation}
\label{sec:implementation}

\textbf{Model.} In order to obtain visual affordance maps of the environment, we adopt an auto-encoder architecture based on a fully convolutional U-Net model \cite{Ronneberger2015UNet}. The input $x$ to the model is a 2D point cloud representation of the scene, which is obtained by projecting the depth information from the camera into the world frame, using the camera parameters. The input goes through the encoder obtaining a latent vector, to which we concatenate the action orientation parameters vector $q$. After going through the decoder, where the skip-connections of the U-Net are applied, we obtain a map of the same size as the input, which we pass through a sigmoid layer to obtain outputs in $[0,1]$. At each point, the output represents the parameter of a Bernoulli distribution that gives the probability of successfully applying the affordance primitive with the corresponding pixel position $p$ and orientation $q$.   
For computing information gain estimates for affordance discovery, we instantiate a lightweight ensemble, where the ensemble shares the encoder but multiple decoders. We found that sharing the encoder overall simplifies training (higher accuracy) while still allowing us to compute the information gain estimates with reduced computation overhead \cite{laurent2023PackedEnsembles}.

We refer to $\theta$ as the set of parameters of one member of the ensemble, which represents our discrete space of parameters $\Theta$. Each member of the ensemble is trained by minimizing a binary cross-entropy loss, where the targets are the success labels of the actions.
\begin{align}
\label{eq:loss}
    \mathcal{L(\theta)} = \sum_n &-\log p(b=\textrm{success}|x_n,a_n,\theta)\cdot b_n \\  
    & -\log p(b=\textrm{fail}|x_n,a_n,\theta)\cdot (1-b_n) \notag
\end{align}
Following previous work \cite{Lakshminarayanan2017DeepEnsemble, Ovadia2019TrustModelUncertainty}, we do not employ bootstrapping, as it can degrade the single model's performance.

\begin{figure*}[t]
\begin{minipage}[b]{0.42\textwidth}
    \centering
    \includegraphics[width=\textwidth]{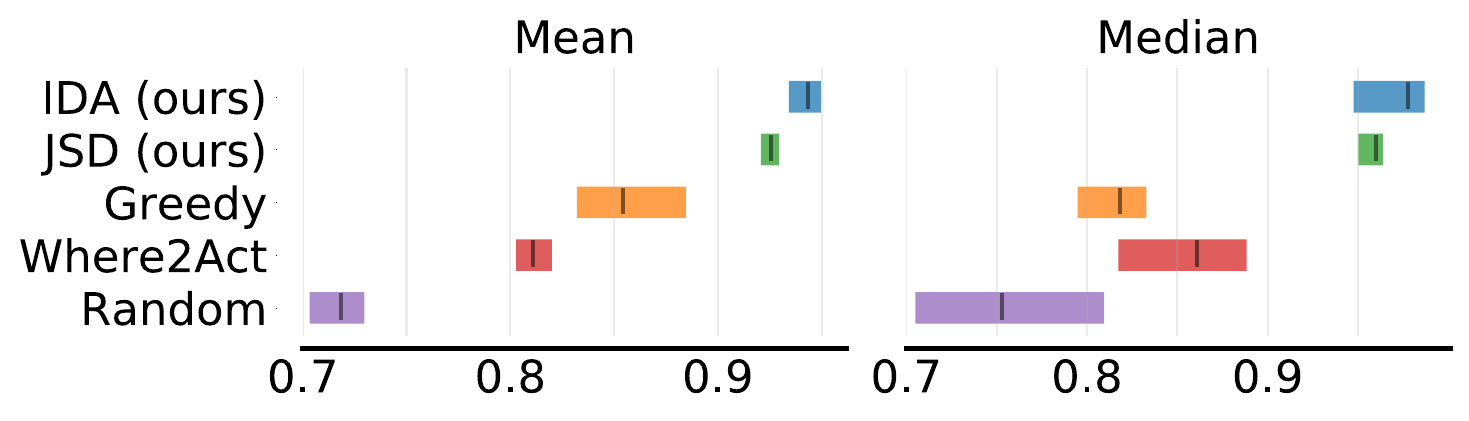}
    \captionof{figure}{\textbf{ManiSkill2 performance.} Affordance success aggregated across ManiSkill2 tasks and runs. }
    \label{fig:main_rliable}
\end{minipage}
\hfill
\begin{minipage}[b]{0.565\textwidth}
\centering
\centering
\resizebox{\columnwidth}{!}{%
\begin{tabular}{c|cccccc}
\hline
  & \textcolor{tblue}{IDA (ours)} & \textcolor{tgreen}{JSD (ours)} & \textcolor{torange}{Greedy} & \textcolor{tred}{Where2Act} & \textcolor{tpurple}{Random} & \textcolor{tgray}{C-GraspNet} \\ \hline
Grasp Cube & \textbf{0.99\stddev{±0.01}} & 0.96\stddev{±0.02} & \textbf{0.97\stddev{±0.04}} & 0.95\stddev{±0.01} & 0.81\stddev{±0.04} & 0.99 \\
Grasp YCB & \textbf{0.91\stddev{±0.02}} & 0.85\stddev{±0.03} & 0.82\stddev{±0.03} & 0.66\stddev{±0.05} & 0.50\stddev{±0.05} & 0.43 \\
Grasp EGAD & \textbf{0.86\stddev{±0.02}} & \textbf{0.83\stddev{±0.03}} & 0.76\stddev{±0.11} & 0.70\stddev{±0.04} & 0.57\stddev{±0.08} & 0.52 \\
Stack Cube & \textbf{0.99\stddev{±0.01}} & \textbf{0.99\stddev{±0.01}} & \textbf{0.99\stddev{±0.01}} & 0.88\stddev{±0.03} & 0.96\stddev{±0.04} & - \\
Open Drawer & \textbf{0.98\stddev{±0.05}} & \textbf{1.00\stddev{±0.00}} & 0.73\stddev{±0.14} & 0.86\stddev{±0.06} & 0.75\stddev{±0.09} & - \\
\end{tabular}%
}
    \captionof{table}{\textbf{Performance per task.} Dissecting performance per task on the ManiSkill2 tasks (mean ± standard deviation over 5+ seeds).}
    \label{tab:main}
\end{minipage}
\end{figure*}

\textbf{Information-driven sampling.} 
In our model, a set of model parameters $\theta$ can be obtained by uniformly sampling models from the set of ensemble parameters $\Theta$, so that the information radius can be rewritten as:
\begin{align} \label{eq:jsd_ensemble}
    I(x,a) &= JSD(p(b|x,a, \theta)|\theta \sim \Theta) = \\
        &= H\left(\frac{1}{N} \sum_{i=1}^N p(b|x,a, \theta_i)\right) -  \frac{1}{N} \sum_{i=1}^N H(p(b|x,a, \theta_j)). \notag
\end{align}
Crucially, the entropy computation is done per pixel, with minimal computation burden.

In order to foster exploration further, we opt for sampling the expected reward $\hat{r}(x,a)$ using a randomly sampled model from the ensemble. This technique, which can be interpreted as a form of Thompson sampling \cite{Russo2020ThompsonSampling} with ensembles \cite{Lu2023Ensemble, Osband2016BootDQN}, can particularly impact the initial phase of learning, when the differently initialized models tend to focus on different areas of the scene.

\textbf{Robust evaluation.} After the training stage, for evaluation, we can use the information gain as a measure of the amount of missing information with respect to a certain interaction. This measure can be used to select more robust actions, following a strategy of ``pessimism" \cite{Kumar2020CQL, Shi2022PessimisticQ}.

The information $I(x,a)$, as described in Equation \ref{eq:jsd}, can be subtracted from the expected rewards, obtaining a more conservative estimate of the action's value:
\begin{align}
    \begin{split}
\label{eq:robust}
    \arg \max_{a} [ \hat{r}(x,a) - c_\textrm{eval} \cdot I(x,a) ]
    \end{split}
\end{align}
where the coefficient $c_\textrm{eval} \geq 0$ controls how conservative the actions should be. Conversely with respect to Equation \ref{eq:ucb}, this equation satisfies a lower confidence bound with respect to the expected reward, which entails pessimism in the face of lack of information \cite{Rashidinejad2023Pessimims}. 
During the evaluation process, in order to sample our most robust expected reward predictions, we use the mean of the affordance probability maps, obtained by averaging the probability outputs of all models in the ensemble, i.e.  $\hat{r}(x,a)=\frac{1}{N} \sum_{i=1}^N p(b|x,a, \theta_i)$.


\section{Experiments}

With our empirical evaluation, we aim to answer the following questions: \textit{i)} what are the effects of adopting \OurMethod{} for the affordance discovery problem in terms of data efficiency and final performance, \textit{ii)} what is the impact of employing the JSD term for sampling informative actions. We perform experiments both in simulated and real-world settings. We also perform some ablations on the components of \OurMethod{}. 

\subsection{Simulation experiments}

To test our approach in an interactive simulation setting, we adopt environments from the ManiSkill2 benchmark \cite{Gu2023Maniskill2}. The scene is recorded by an external RGBD camera that points at the scene, showing the robot and workspace in front of it. 
During training the agent alternates between environment interactions, where it tests the action primitive in an \textit{interaction}, and updates the affordance model after each interaction. An interaction roughly requires $100$ simulation steps. 
Every $n$ interactions,  where {$n \in$ \{ 50, 250, 500, 2500, 10000 \}}, we evaluate the agent's performance. 


We consider the following set of primitives and corresponding affordances:
\begin{itemize}
    \item \textit{Grasping:} success is achieved if the object can be lifted without falling out of the gripper.
    \item \textit{Stacking:} success is achieved if the object stably remains on the other object after the gripper releases the held object and withdraws.
    \item \textit{Opening:} success is achieved if a part of the object is pulled and that classifies as an opening part, e.g., the drawer in a cabinet.
\end{itemize}
For all primitives, we adopt a discrete set of orientations with 6-DoF for grasping (429 angles) and 3-DoF for stacking and opening (8 angles). Actions are executed by performing (linear) motion planning. Further details are given in Appendix.

\textbf{Results.} We present evaluation results, in terms of affordance success rate, for \OurMethod{} and the following baselines:
\begin{itemize}
    \item \textit{JSD (ours)}: this method only samples actions according to information gain during training, and uses the ensemble predictions for evaluation. It represents an ablation of IDA, using no UCB and no robustness during evaluation and showcases the performance of the information gain objective;
    \item \textit{Where2Act}: this baseline mimics the Where2Act sampling strategy \cite{Mo2021Where2Act}, while employing our U-net based architecture. The agent randomly samples actions during the first 5000 training steps, while during the latter 5000 training steps, it samples from the softmax probability distribution over all possible actions, essentially doing a form of Boltzmann exploration \cite{Cesabianchi2017BoltzmannDoneRight}. Note that in our experiments we use a mounted robot arm, as opposed to the flying gripper adopted in the original work.
    \item \textit{Greedy}: this method uses a U-Net and selects the highest probability affordance actions, both for training and evaluation;
    \item \textit{Random}: the agent randomly samples affordance actions during training and selects the highest probability affordance action for evaluation, like Greedy.
\end{itemize}
For the grasping tasks, we also compare to \textit{Contact-GraspNet} \cite{Sundermeyer2021ContactGraspnet}, a grasping approach that has been trained on 17.7 million simulated grasps across 8872 meshes, for which we use the original open-source pre-trained models and code. 

In Figure \ref{fig:main_rliable}, we present bootstrapped statistics and confidence intervals for the mean and the median performance after 10k steps aggregated across all tasks and runs (5+ seeds per method and task), using 50k repetitions (see \cite{Agarwal2021RLiable}). We observe that \OurMethod{} reaches the highest mean, with high confidence, and the highest median, followed by JSD (ours). Both our methods largely outperform all the other baselines. 

In Table \ref{tab:main}, we present the mean and standard deviation statistics per task. We observe that on the most difficult tasks, Grasp EGAD (large number of variations) and Open Drawer (harder exploration), \OurMethod{} and JSD have the largest advantage, corroborating the idea that information-driven affordance discovery leads to higher final performance. Compared with Contact-GraspNet in the grasping tasks, we also observe that all interactive learning methods, except for Random, eventually outperform pre-training on large synthetic datasets. 

\begin{figure}[t]
\includegraphics[width=0.9\linewidth]{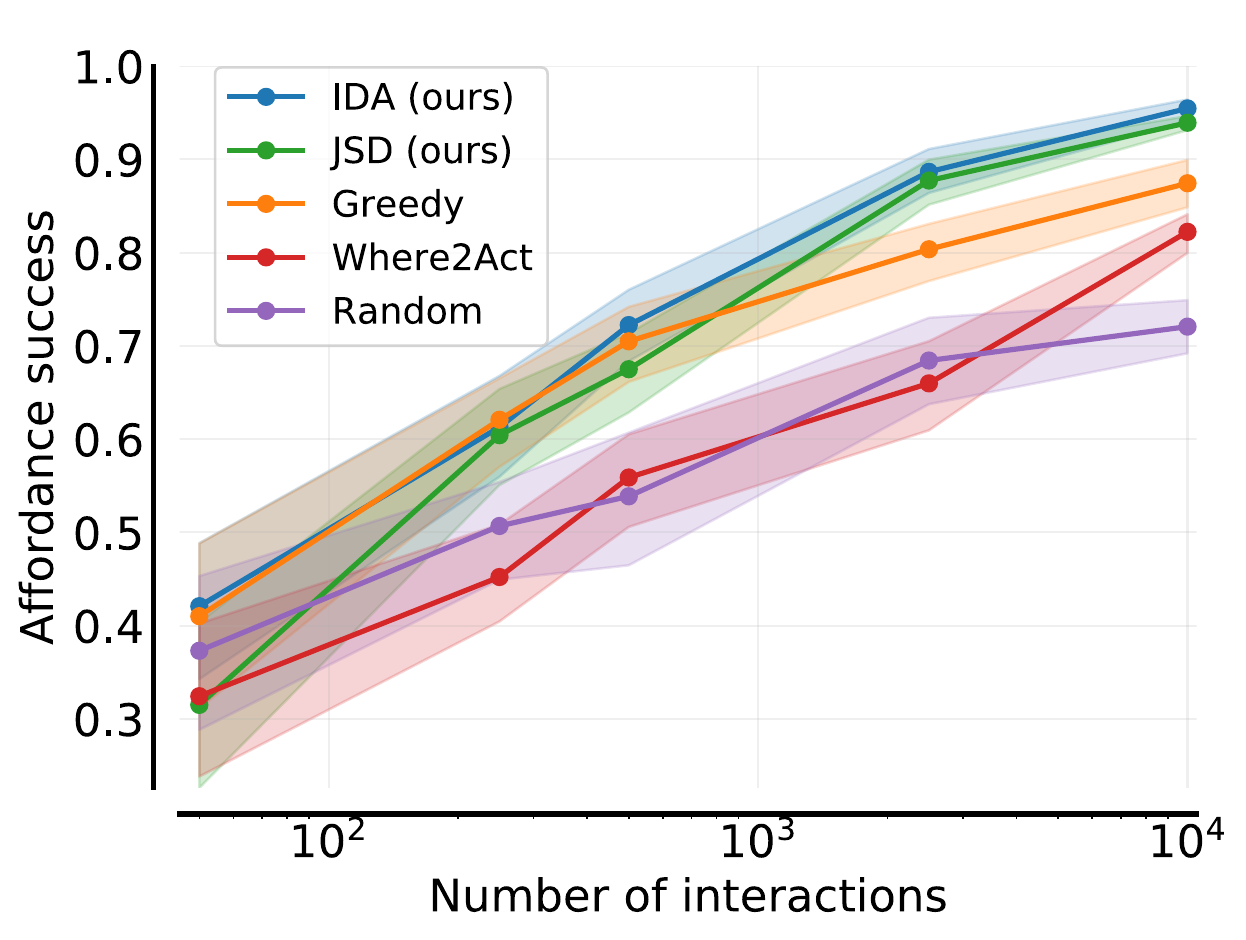}
\caption{\textbf{Performance over time.} 
The affordance success rate in the evaluation stage increases over the number of interactions, averaged over all tasks (5+ seeds per task).} 
\label{fig:main_time}
\end{figure}

Finally, in Figure \ref{fig:main_time}, we compare performance over time aggregated across all tasks and runs, using bootstrapped statistics and confidence intervals with respect to the mean \cite{Agarwal2021RLiable}. We observe that while JSD's final performance is close to \OurMethod{}'s one, the Greedy approach performs better in the early stages of learning. This shows that the reward term, which we incorporate in IDA using a UCB, can be useful when sampling actions. We speculate the reason for this being that during the first stages of learning, it allows the agent to learn and consolidate knowledge about easy affordances, before moving to more complex ones.    

\begin{figure}[b]
\includegraphics[width=0.9\linewidth]{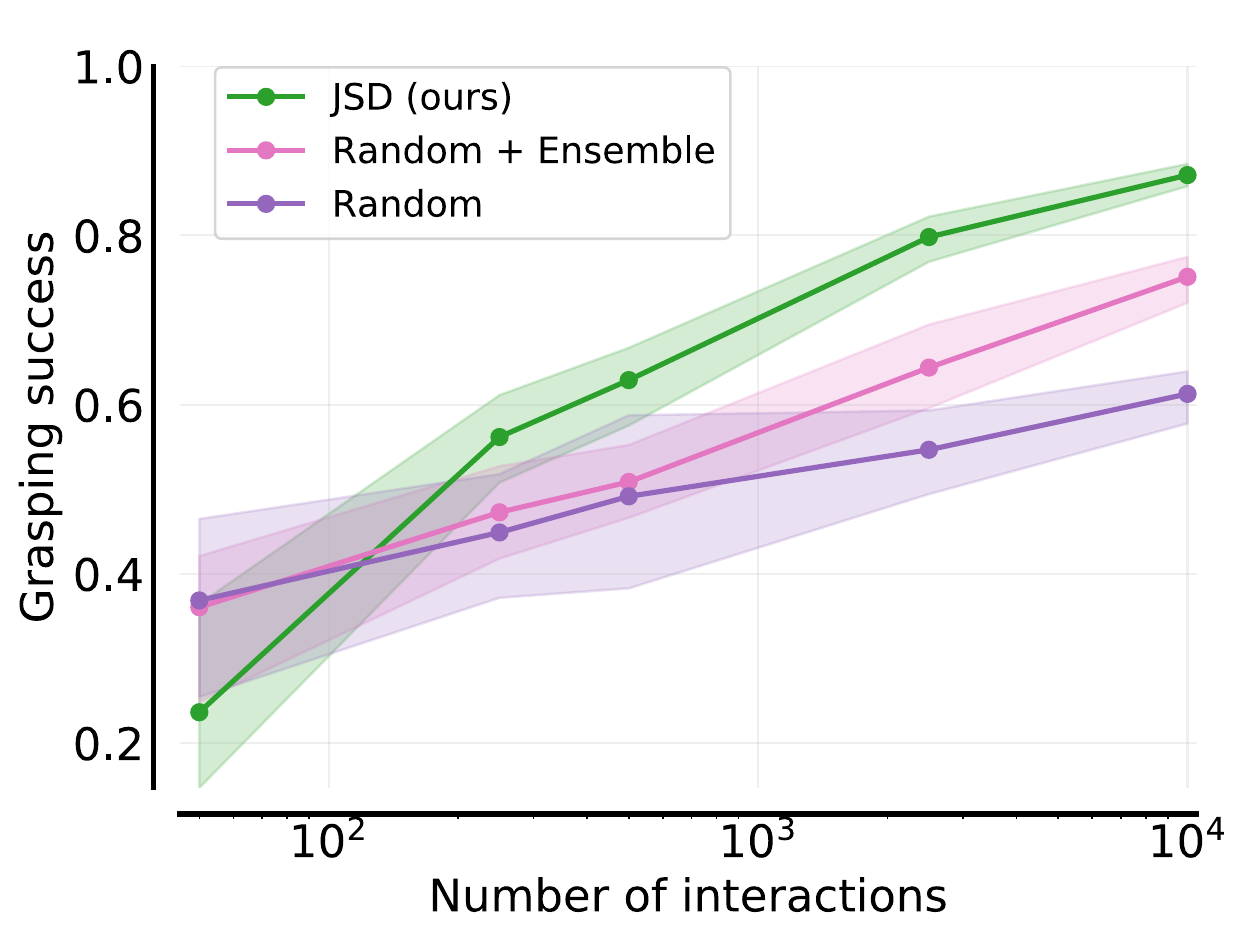}
\caption{\textbf{Reward-free ablation.} 
Comparing reward-free affordance discovery methods over time. (5+ seeds). } 
\label{fig:expl_time}
\end{figure}

\textbf{Reward-free ablation.} The objective of IDA is made of two components: a reward-driven term and an information-driven term. \textit{What happens when we ignore the reward-driven term?} In that case, the other approaches (including Where2Act) rely on random sampling, while \OurMethod{} relies on the JSD strategy. We more detailedly compare these two strategies over time on the three ManiSkill grasping tasks (Cube, YCB and EGAD), along with a \textit{Random + Ensemble} strategy, which aims to reduce the gap between the two methods, by learning an ensemble model for evaluation. In Figure \ref{fig:expl_time}, we observe that although the performance of Random increases, during the later stages, when employing an ensemble for evaluation, the JSD sampling strategy largely remains the most effective one. 

Interestingly, JSD's performance is lower at the first evaluation (after only 50 interactions). This is probably due to the fact that random samples possess greater diversity at the beginning of training when the JSD information estimates, and thus action sampling, are not yet well-directed. As we show in Table \ref{tab:initial}, the use of rewards and of ensemble sampling in \OurMethod{} both help in mitigating the issue. 

\begin{table}[h]
\centering
\begin{tabular}{ccc}
\hline
IDA &   \renewcommand{\arraystretch}{0} \begin{tabular}[c]{@{}c@{}}IDA\\ \textcolor{tgray}{\footnotesize{(no ens. sampl.)}} \end{tabular} \renewcommand{\arraystretch}{1} & JSD  \\ \hline
\textbf{0.45} & 0.39 & 0.25 \\
\end{tabular}%
    \caption{Grasping success (avg) after 50 interactions.}
    \label{tab:initial}
\end{table}

\subsection{Real-world experiments} 

To confirm that our simulation results hold in the real world, we deployed and tested our method on a grasping task using a \xArm\ with UFACTORY gripper. 

We studied the grasping primitive using a set of four toy objects of varying size. 
The objects were presented repeatedly in order during both training and evaluation and manually placed at random locations and orientations in the workspace of the robot arm before each grasp attempt. 

Differently from the simulation setup, the grasping angle was restricted to vertical grasps with eight discrete rotations (3 DoF) and a wrist-mounted RGBD camera is used to obtain the inputs. This also shows our method works indipendently of the viewpoint.
The affordance model is always trained from random initialization (no prior knowledge) and evaluated at 25, 50, 100, 250 episodes. Evaluation is done for 20 iterations, i.e., each of the four objects is presented five times per evaluation round. The entire experiment, including training and evaluation, takes around 90 minutes. 

\begin{figure}[t]
    \centering
    \includegraphics[width=0.8\linewidth]{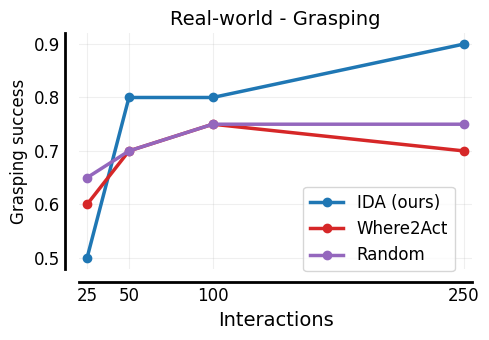}
    \includegraphics[width=0.3475\linewidth]{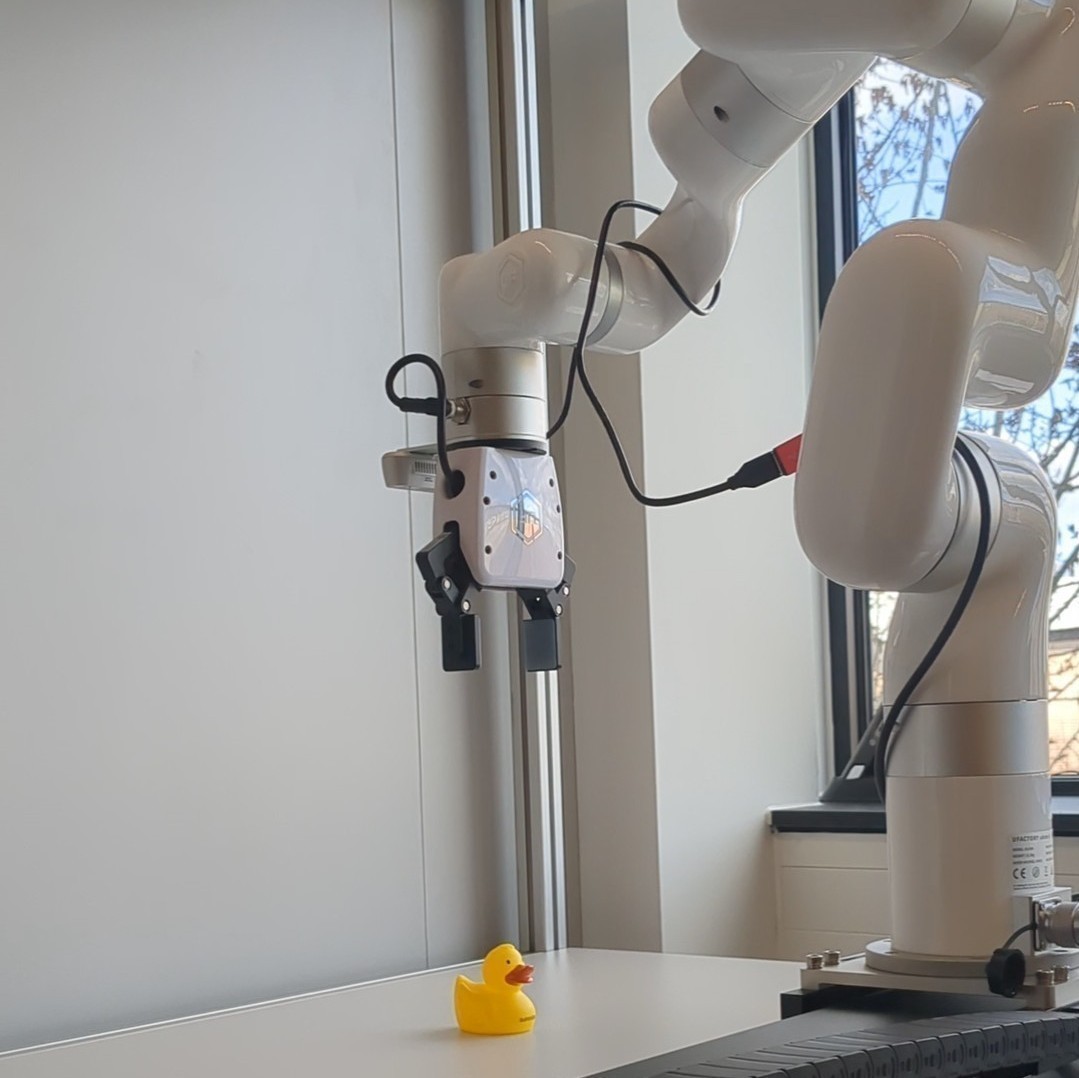}
    \hfill
    \includegraphics[width=0.55\linewidth]{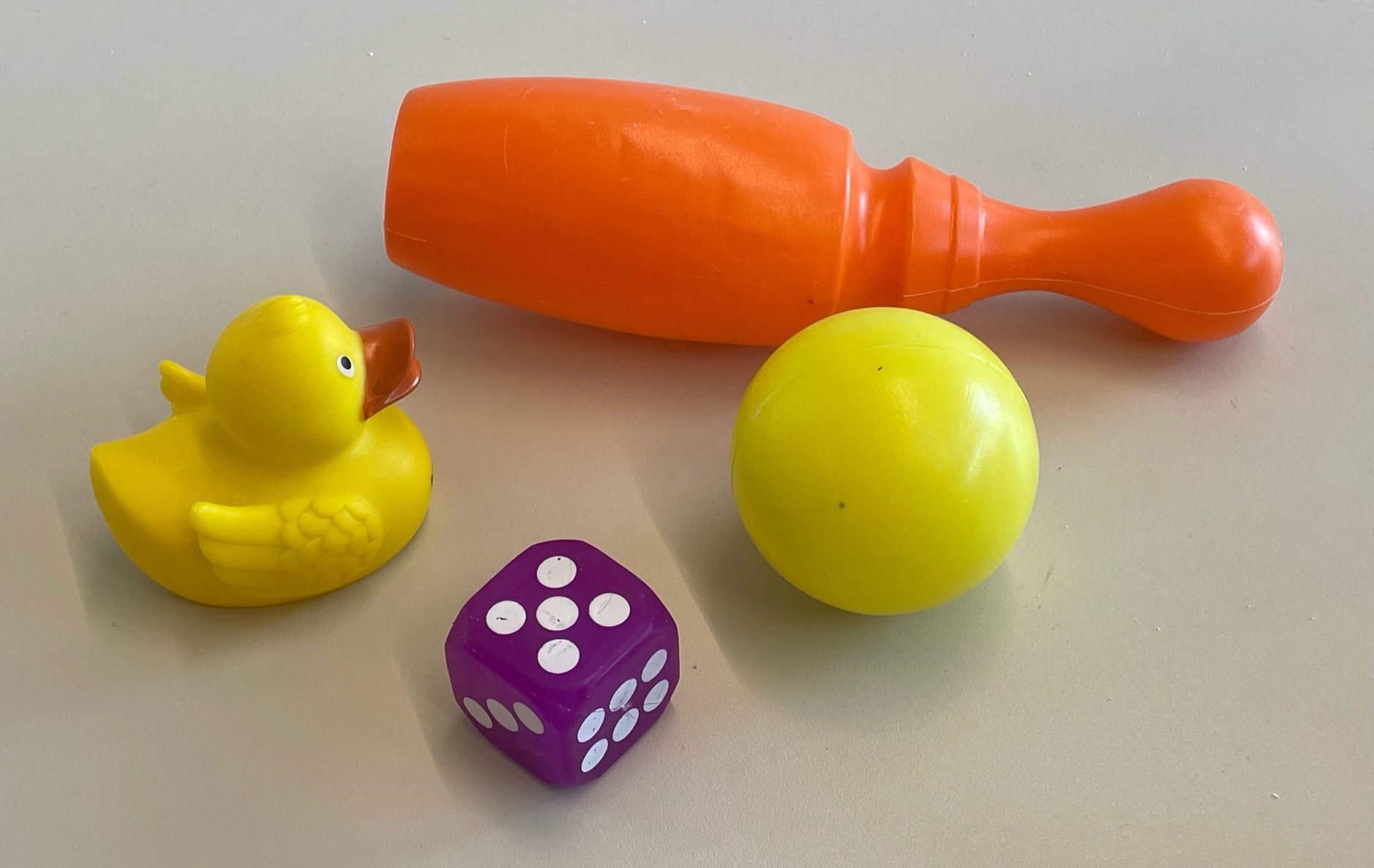}        
\caption{\textbf{Real-world results and setup.} \OurMethod{} learns to grasp objects faster than other approaches, achieving up to 90\% grasping success, on a \xArm \  platform. }
\label{fig:real_world_all}
\end{figure}

\begin{figure}[t]
    \centering
    \includegraphics[width=\linewidth]{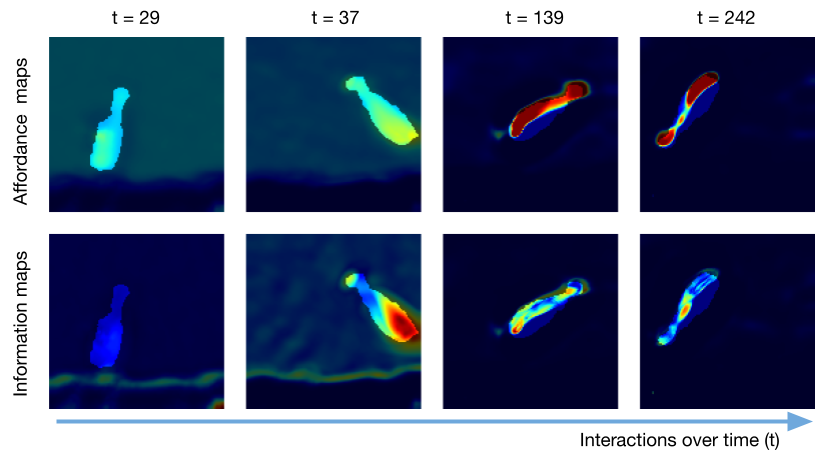}
    \caption{
\textbf{Real-world affordance and information maps.} Affordance and information maps, showing for each pixel the highest value across all possible gripper orientations. 
}
\label{fig:vis_afford_real}
\end{figure}

\textbf{Results.} In Figure~\ref{fig:real_world_all} we compare \OurMethod{} with two baselines: the Where2Act and the Random action sampling strategies. We observe the real-world results for grasping closely match the simulation results, as our agent efficiently learns to grasp the objects. The final performance obtained is 90\% grasping success, which is considerably higher than the baselines. Nonetheless, we experienced lower performance than Random and Where2Act (also using random sampling during the initial steps) after only 25 interactions. This result reflects what we found in the reward-free ablation experiments, and how to further mitigate this early-learning effect will be the object of future studies. 

\textbf{Visualizations.} To provide additional insights into how \OurMethod{} learns visual affordances and estimates information gain over time, in Figure \ref{fig:vis_afford_real}, we show how the affordance and information maps from our method evolve over time while learning to grasp (a bowling pin) on the real robot arm. We observe that the affordance probabilities are uniformly distributed at the beginning of the training ($t=29$). Later, the information maps recommend exploring grasping point towards the edge of the object ($t=37$, $t=139$), leading the agent to eventually learn that areas close to the edges are easier to grasp ($t=242$) as they have a less slippery surface. 


\section{Conclusion} 
\label{sec:conclusion}

We presented \OurMethod{}, a method that fosters the discovery of affordances for robotics manipulation. \OurMethod{} showcases strong performances across several tasks in ManiSkill2 and the ability to quickly learn to grasp objects in the real world. We empirically showed the importance of well-directed action sampling, to achieve higher affordance success and we analyzed several components of our method. One limitation of the approach tested is that it relies on motion planning for precise affordance execution. While this facilitates exploration, especially in the early stages of learning, as the actions executed by the agent are more stable and reliable, the issue should be addressed in future work, aiming to provide more adaptive policies, e.g. using reinforcement learning. We also aim to extend our work towards the development of an end-to-end system able to solve longer horizon tasks, potentially instantiating a hierarchical controller on top of the possible affordance actions \cite{Hafner2022Director} or employing large language models to decide which affordances should be executed towards the solution of tasks \cite{Ahn2022SayCan, Lin2023Text2motion}. 



\section*{Appendix}

\subsection{Information radius derivation}
\label{app:derivation}

As per Equation \ref{eq:info}, we are interested in measuring the expected information given by a certain state-action:
\begin{align}
    I(x,a) 
    = \sum_{b}p(b|x,a)\KL{}[p(\Theta|b,x,a) \Vert p(\Theta)], \notag
\end{align}
where we can use the KL definition to obtain:
\begin{align}
    I(x,a) &= \sum_{b}p(b|x,a)\sum_{\theta}p(\theta|b,x,a)\log\frac{p(\theta|b,x,a)}{p(\theta)}. \notag
\end{align}
Using the Bayes rule ($p(\theta|b,x,a)=\frac{p(b|x,a,\theta)p(\theta|x,a)}{p(b|x,a)}$) and the observation $p(\theta|x,a)=p(\theta)$ (as an incomplete transition provides no information about the affordance model) we get: 
\begin{align}
    I(x,a) 
  &= \sum_{b}\sum_{\theta}p(b|x,a,\theta)p(\theta)\log\frac{p(b|x,a,\theta)}{p(b|x,a)}.  \notag
\end{align}
By splitting the logarithm and applying the law of total probability we derive:
\begin{align}
 I(x,a) &= \sum_{\theta}p(\theta)\sum_{b}\textcolor{tgreen}{p(b|x,a,\theta)}\textcolor{torange}{\log p(b|x,a,\theta)} \notag \\
 & -\sum_{b}\textcolor{tgreen}{\sum_{\theta}p(\theta)p(b|x,a,\theta)}\textcolor{torange}{\log(\sum_{\theta}{p(\theta)p(b|x,a,\theta)})} \notag,
\end{align}
that given entropy $H(Y)=-\sum_y \textcolor{tgreen}{p(y)}\textcolor{torange}{\log p(y)}$ becomes:
\begin{align}
 I(x,a) &= -\mathbb{E}_{p(\theta)}[H(p(b|x,a,\theta))] + H(\mathbb{E}_{p(\theta)}[p(b|x,a,\theta)]) \notag
\end{align}
that is the JSD we present in Equation \ref{eq:jsd}.

\subsection{Training details}

\textbf{Model.} The ensemble model is a U-Net-like model made of a shared encoder and multiple decoders. The encoder has a convolutional block with 8 filters, kernel size 3 and stride 1, followed by two blocks with 16 filters, kernel size 5 and stride 2. The decoder follows the same structure reversed, but replacing stride with bilinear upsampling layers and applying skip connections. The inputs' resolution for the model is 128x128. The number of networks in the ensemble is 5. The model is updated 5 times after each interaction, using ADAM with a learning rate of 3$\cdot10^{-4}$. The batch size used is 256. To start filling the replay buffer with experience, 10 random actions are sampled at the beginning of training.

\textbf{Action sampling.} To avoid sampling invalid points
we filter out some regions of the environment from the output of the model, including points that are behind and above the robot's base and points that are on the surface of the workspace or below. 
Before summing the information gain values with the predicted affordance success reward, we normalize the information gain maps to have values in [0,1]. As default values, we use $c_\textrm{expl}=0.3$ and $c_\textrm{eval}=0.1$.


\subsection{Evaluation details}

\begin{itemize}
    \item \textit{Grasp Cube}: 100 randomized positions.
    \item  \textit{Grasp EGAD}: 1 interaction for each of the 1600 objects.
    \item \textit{Grasp YCB}: 5 interactions for each of the 74 objects. 
    \item \textit{Stack Cube}: 100 randomized interactions
    \item \textit{Open Drawer}: 5 interactions for each cabinet (6 models) 
\end{itemize}


\bibliographystyle{IEEEtran}
\bibliography{IEEEabrv,references}


\end{document}